\documentclass[runningheads]{llncs}

\usepackage{eccv}

\usepackage{eccvabbrv}

\usepackage{graphicx}
\usepackage{booktabs}

\usepackage[accsupp]{axessibility}  

\usepackage{hyperref}

\usepackage{xcolor}
\usepackage{multirow}
\usepackage{makecell}
\usepackage[symbol]{footmisc}

\usepackage{orcidlink}

\begin{document}

\title{Unleashing Text-to-Image Diffusion Prior for Zero-Shot Image Captioning} 

\author{Jianjie Luo\inst{1} \and
Jingwen Chen\inst{2} \and
Yehao Li\inst{2} \and
Yingwei Pan\inst{2} \and
Jianlin Feng\inst{1}$^\dagger$ \and
Hongyang Chao\inst{1} \and
Ting Yao\inst{2}$^\dagger$}

\authorrunning{J.~Luo et al.}

\institute{Sun Yat-sen University, Guangzhou, China \and
HiDream.ai Inc.\\
\email{luojj26@mail3.sysu.edu.cn}, 
\email{\{fengjlin, isschhy\}@mail.sysu.edu.cn} \\
\email{\{chenjingwen, liyehao, pandy, tiyao\}@hidream.ai}
}

\maketitle


\begin{abstract}
\footnotetext[2]{ J. Feng and T. Yao are the corresponding authors.}

Recently, zero-shot image captioning has gained increasing attention, where only text data is available for training. The remarkable progress in text-to-image diffusion model presents the potential to resolve this task by employing synthetic image-caption pairs generated by this pre-trained prior. Nonetheless, the defective details in the salient regions of the synthetic images introduce semantic misalignment between the synthetic image and text, leading to compromised results. To address this challenge, we propose a novel \textbf{P}atch-wise \textbf{C}ross-modal feature \textbf{M}ix-up (PCM) mechanism to adaptively mitigate the unfaithful contents in a fine-grained manner during training, which can be integrated into most of encoder-decoder frameworks, introducing our PCM-Net. Specifically, for each input image, salient visual concepts in the image are first detected considering the image-text similarity in CLIP space. Next, the patch-wise visual features of the input image are selectively fused with the textual features of the salient visual concepts, leading to a mixed-up feature map with less defective content. Finally, a visual-semantic encoder is exploited to refine the derived feature map, which is further incorporated into the sentence decoder for caption generation. Additionally, to facilitate the model training with synthetic data, a novel CLIP-weighted cross-entropy loss is devised to prioritize the high-quality image-text pairs over the low-quality counterparts. Extensive experiments on MSCOCO and Flickr30k datasets demonstrate the superiority of our PCM-Net compared with state-of-the-art VLMs-based approaches. It is noteworthy that our PCM-Net ranks first in both in-domain and cross-domain zero-shot image captioning. The synthetic dataset SynthImgCap and code are available at \url{https://jianjieluo.github.io/SynthImgCap}.

\end{abstract}


\section{Introduction}
\label{sec:intro}

\begin{figure}[!tb]
    \centering {\includegraphics[width=0.99\textwidth]{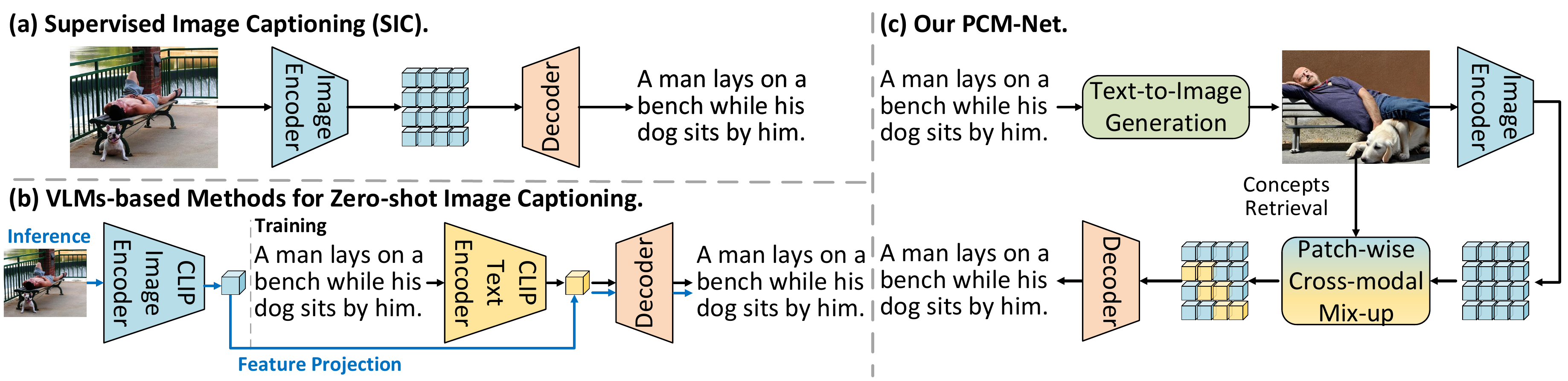}}
    \caption{Training paradigms for image captioning: (a) Training with well-aligned image-sentence pairs for Supervised Image Captioning (SIC); (b) Training with text-only data in VLMs-based model for Zero-shot Image Captioning (ZIC); (c) Training with synthetic image-text pairs in our PCM-Net for ZIC.}
    \label{fig:intro}
\end{figure}

Image captioning aims to describe the content of an image with natural language. 
The conventional practice to resolve this task is to train an encoder-decoder model in an end-to-end manner with well-aligned pairs of images and texts, widely known as Supervised Image Captioning (SIC) (Figure \ref{fig:intro} (a)). More advanced methods are proposed \cite{pan2016jointly,yao2017incorporating,anderson2017bottom,li2019pointing,huang2019attentio,luo2021coco,li2021scheduled,li2022comprehending,luo2023semantic} to better capture the correlation between visual and linguistic elements. However, the extensive labor required for the data collection makes it difficult for SIC model to scale up.

To address this challenge, Zero-shot Image Captioning (ZIC) is introduced and manages to learn the image-to-text mappings without relying on annotated data, which has attracted great interest recently. The popular recipe for ZIC is to bridge vision and language through an intermediate latent representation using unpaired images and sentences \cite{feng2019unsupervised,gu2019unpaired}. Another type of VLMs-based solution (Figure \ref{fig:intro} (b)) for ZIC involves developing an image captioning model with high-quality text-only data, where the pre-trained cross-modal prior from large-scale vision-language models (VLMs) like CLIP \cite{radford2021learning} is leveraged to align the text and image. In VLMs-based approaches, the captioning model is trained to generate caption with the corresponding textual feature in CLIP space, which will be replaced with the visual feature of the input image during inference. It is assumed that the textual and visual features sharing similar semantics should be close in CLIP space. However, this assumption may not always hold due to the inherent modality gap \cite{liang2022mind}, leading to \emph{discrepancy between training and inference}.

Motivated by the powerful capability of text-to-image diffusion model \cite{ramesh2022hierarchical,saharia2022photorealistic,rombach2022high} in producing images conditioning on text prompts, we propose to generate synthetic images for the text data, and train the captioning model with the obtained synthetic image-text pair to mitigate the above issue. Nevertheless, the presence of flawed or unfaithful details in the salient regions of the synthetic images enlarges not only the distribution discrepancy between the synthetic and real images but also the semantic gap between the synthetic image and the text, thereby hindering the learning of visual-semantic alignment. Since generating high-fidelity images usually requires rigorous prompt engineering, our work primarily aims to automatically mitigate the unfaithful contents in the synthetic images during training. To achieve this goal, a novel patch-wise cross-modal feature mix-up (PCM) mechanism is devised in this paper, which can be integrated into most of encoder-decoder frameworks, introducing our PCM-Net (Figure \ref{fig:intro} (c)). Specifically, given an input image, a set of salient visual concepts is constructed by performing zero-shot entity classification of images in CLIP space. Then, the fine-grained patch features of the synthetic image are mixed up with the textual features of the salient visual concepts depending on patch-wise cross-modal similarity, which effectively reduces the flawed or unfaithful details in the synthetic images. Finally, the derived features are passed through an encoder-decoder network for caption generation. Compared to the former VLMs-based approaches which rely on the global feature of CLIP, our PCM-Net mitigates the modality gap in pixel space by using synthetic images for training, where fine-grained visual features are available to boost the visual-semantic alignment. Additionally, we propose a new CLIP-weighted cross-entropy loss to facilitate the robust training of the captioning model with noisy synthetic data by adaptively re-weighting the loss for each word according to the predicted word distribution and the semantic relevance between the text and the synthetic image.

To summarize, our contributions are as follows:
(\textbf{I}) We propose leveraging the powerful text-to-image diffusion model to generate synthetic images for text-only data, and further train the captioning model with these synthetic pairs, which mitigates the discrepancy between training and inference for zero-shot image captioning. (\textbf{II}) We propose a patch-wise cross-modal feature mix-up (PCM) mechanism to close the gap between synthetic and real images by automatically mitigating the defective contents in the synthetic images. (\textbf{III}) We propose a novel CLIP-weighted cross-entropy loss (CXE) to improve the training of captioning model with the noisy synthetic data. (\textbf{IV}) Extensive experiments conducted on MSCOCO and Flickr30k demonstrate our PCM-Net's effectiveness.


\section{Related Work}
\label{sec:related_work}

\subsection{Supervised Image Captioning}
Image captioning, a fundamental task in the vision-language domain, aims to describe semantic content within an image using natural language. Conventional image captioning methods train an encoder-decoder neural network on well-matched image-sentence pairs in a supervised manner. Early attempts \cite{karpathy2015deep,Vinyals14,donahue2015long,Xu:ICML15} leverage CNNs to encode visual content and RNNs to decode sentences. Later researches focus on enriching the visual feature representation by incorporating semantic attributes \cite{You:CVPR16,yao2017boosting} or object region features \cite{anderson2017bottom}. The performances of supervised image captioning are further boosted by modeling visual relationships in the image \cite{yao2018exploring,Yang:CVPR19,yao2019hierarchy}. Recently, inspired by the success of the Transformer architecture \cite{vaswani2017attention}, numerous Transformer-based image captioning models have emerged \cite{herdade2019image,cornia2020meshed,pan2020x,yang2021auto,luo2023semantic}. Despite the significant progress in supervised learning tasks, the collection of annotated training datasets is labor-intensive and time-consuming, which restricts the scalability of the captioning model.

\subsection{Zero-shot Image Captioning}
Zero-shot image captioning aims to learn the captioning model without human-annotated data. Recently, the vision-language pre-training models (VLMs) trained on large-scale image-text pairs crawled from the web have demonstrated great zero-shot capability in image captioning. For example, SimVLM \cite{wang2021simvlm} pre-trains the VLM with a single prefix language modeling objective and infers the image caption without fine-tuning. BLIP-2 \cite{li2023blip} only pre-trains a Querying Transformer to bridge the modality gap between the pre-trained image encoder and the large language model to further boost zero-shot image captioning. ZeroCap \cite{tewel2022zerocap} instead steers GPT-2 to generate sentences related to the visual content under the guidance of CLIP \cite{radford2021learning} without further training. However, due to the noise in the web data, these VLMs-based approaches struggle to generate captions with correct grammar without fine-tuning on paired data. 

Therefore, another approach for ZIC is to build an image captioning model based on high-quality text-only data. The prevalent methods \cite{su2022language,nukrai2022text,li2023decap} leverage the pre-trained cross-modal prior of the VLMs to facilitate the learning of visual-semantic alignment for zero-shot image captioning. For instance, MAGIC \cite{su2022language} fine-tunes GPT-2 on the training corpus, and modulates the probability of each candidate word during decoding with the cross-modal similarity in CLIP space. Furthermore, several works \cite{nukrai2022text,li2023decap,fei2023transferable,yu2023cgt} propose to train the captioning model with the textual features of the target caption as input at training stage, and replace it with the visual feature of the input image during inference. The pioneering CapDec \cite{nukrai2022text} applies noise injection to input textual features in training to overcome the modality gap \cite{liang2022mind}, while DeCap \cite{li2023decap} projects the visual feature into the textual feature space based on the training corpus at inference for the same purpose. CgT-GAN \cite{yu2023cgt} instead mitigates the modality gap by incorporating real images as input in training via additional reinforcement learning rewarded by cross-modal similarity. ViECap \cite{fei2023transferable} builds a text prompt from visual entities to trigger the transferability of GPT-2. 
Most recently, SynTIC \cite{liu2023improving} employs the Stable Diffusion \cite{rombach2022high} model to generate synthetic images for training, and further minimize the global feature distance between synthetic and real images through contrastive learning techniques.

\textbf{Summary.} 
SynTIC \cite{liu2023improving} is most related to our work, which capitalizes on global features of the synthetic images for ZIC. Going beyond SynTIC, our PCM-Net explores spatial visual features from CLIP and leverages its capability of cross-modal alignment to mitigate flawed content in synthetic images, addressing distribution discrepancies with real images at a fine-grained level. Additionally, we introduce a novel CLIP-weighted cross-entropy loss to improve the robustness and performance of the captioning model when trained on noisy synthetic data.

\section{Method}
\label{sec:method}

\begin{figure*}[!tb]
\centering {\includegraphics[width=1\textwidth]{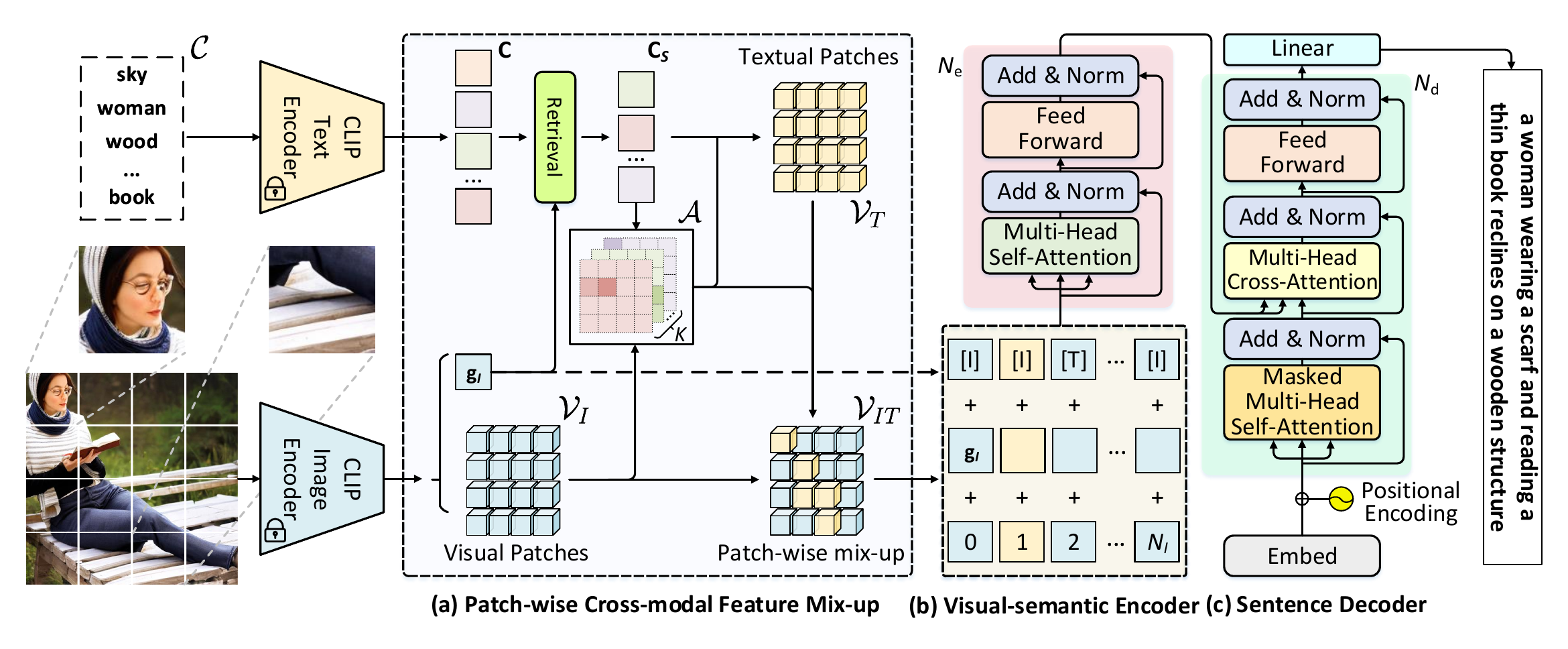}}
\caption{An overview of our proposed PCM-Net. The flawed or unfaithful patches (e.g.,  poor facial details or missing limbs) in the salient regions of synthetic images would be replaced by semantically aligned textual patches in (a) Patch-wise Cross-modal Feature Mix-up. The mixed-up features are further encoded by (b) Visual-semantic Encoder, followed by (c) Sentence Decoder for caption generation.}
\label{fig:framework}
\end{figure*}

Zero-shot image captioning task aims to train an image-to-text captioning model on a text corpus only, without the associated images. Most existing methods learn to generate the caption conditioned on the corresponding textual feature during training, framing it into a task of caption reconstruction. While at inference stage, the textual feature is simply replaced with the global visual feature of the input image within the same multi-modal feature space. As a result, semantic misalignment arises during inference due to the inherent modality gap \cite{liang2022mind} between the caption and the image in the multi-modal space, leading to suboptimal results. To address this issue, we propose leveraging synthetic image-caption pairs generated by text-to-image diffusion models \cite{rombach2022high,ho2020ddpm} for training instead, unleashing the pre-trained diffusion prior for ZIC. However, the distribution discrepancy between the synthetic and real images presents a challenge for training the captioning model. In this paper, we devise a novel Patch-wise Cross-modal Feature Mix-up mechanism to bridge this gap and seamlessly incorporate it into an encoder-decoder captioning framework, introducing our proposed PCM-Net. The framework of our PCM-Net is illustrated in Figure \ref{fig:framework}. In this section, we first demonstrate the details about how to build the synthetic dataset, followed by the specifics of our PCM-Net. Lastly, the training loss is elaborated.

\subsection{Synthetic Image Generation}
The recent progress in text-to-image diffusion models \cite{rombach2022high,ho2020ddpm} has demonstrated the capability of diffusion models to generate high-fidelity images that are semantically aligned with the given texts. This highlights the promise of employing diffusion models to produce synthetic image-sentence pairs that could benefit the cross-modal vision-language learning in zero-shot image captioning.
Let $\mathcal{S}=\{S_i\}^{N}_{i=1}$ denote a text corpus including $N$ sentences for zero-shot image captioning, where each $S_i$ consists of $N_s$ words. Specifically, an off-the-shelf text-to-image generation model $\mathcal{G}$ (i.e., Stable Diffusion \cite{rombach2022high}) is exploited to generate a synthetic image $I_i$ for each sentence $S_i \in \mathcal{S}$, resulting in a synthetic dataset $\mathcal{D} = \{(I_i, S_i)\}^{N}_{i=1}$. 
Please note that all the experiments are conducted on $\mathcal{D}$.

\subsection{Patch-wise Cross-modal Feature Mix-up}
We have found that straightforwardly exploiting synthetic images for training yields compromised results in our experiments. The reason is that the flawed or unfaithful details in the salient areas of these images significantly enlarge not only the discrepancy between the synthetic and real images but also the semantic gap between the synthetic image and the sentence, hindering the learning of visual-semantic alignment. However, high-fidelity image generation usually requires intricate prompt engineering, and a universal solution for different diffusion models has not been established yet. Therefore, we propose a novel Patch-wise Cross-modal feature Mix-up (PCM) mechanism that automatically mitigates the possible irrelevant or defective contents in the salient areas of synthetic images during training, which avoids the complicated prompt engineering or sophisticated model design for faithful text-to-image generation. Specifically, the visual features of the holistic image $I$ and its local patches are first extracted by the image encoder of CLIP. Let ${\mathbf{g}}_I$ and $\mathcal{V}_I=\{{\mathbf{v}}_j\}^{N_I}_{j=1}$ denote the global image feature and the grid feature map with $N_I$ local patch features, respectively.

\textbf{Salient Visual Concepts Detection.}
To bypass the unfaithful or defective contents in the salient areas of the synthetic images, salient visual concepts of each image are required to be first detected, which is achieved by zero-shot entity classification in CLIP space. Specifically, a visual concept vocabulary $\mathcal{C} = \{c_j\}_{j=1}^{N_c}$ is built from the high-frequency nouns in the training corpus $\mathcal{S}$, where ${N_c}$ is the concept vocabulary size. 
Then, the cosine similarity between the template sentence ``A photo of \{$c_j$\}" for each $c_j \in \mathcal{C}$ and the global visual feature $\mathbf{g}_I$ is calculated to retrieve top-$K$ salient visual concepts $\mathcal{C}_s = \{c_k\}_{k=1}^{K}$. Formally, this process is expressed as:
\begin{equation}
\mathcal{C}_s = \underset{c_j \in \mathcal{C}}{{\arg\max}_K}~CosSim(\mathbf{c}_j, \mathbf{g}_I) = \frac{{\mathbf{g}}_I \cdot {\mathbf{c}}_j} {\| {\mathbf{g}}_I \| \| {\mathbf{c}}_j \|},
\end{equation}
where ${\mathbf{c}}_j$ is the textual feature of the corresponding template sentence. Given $\mathcal{C}_s$, a sequence of textual basis vectors ${\mathbf{C}}_s = \{{\mathbf{c}}_k\}_{k=1}^{K}$ are naturally formed by extracting the corresponding global textual feature of each $c_k$.

\textbf{Patch-wise Cross-modal Feature Mix-up.}
Our proposed PCM aims to automatically exclude the potential irrelevant and flawed parts in the salient regions of the synthetic images during training in a fine-grained manner. To achieve this goal, the original visual feature map $\mathcal{V}_I$ is novelly transformed into a new textual feature map $\mathcal{V}_T$ in our PCM by considering the semantic similarity between each local patch $\mathbf{v}_j$ and all the salient concept features ${\mathbf{C}}_s = \{{\mathbf{c}}_k\}_{k=1}^{K}$. However, the modality gap between $\mathbf{v}_j$ and $\mathbf{c}_k$ inherently exists in CLIP space. Inspired by DenseCLIP \cite{rao2022denseclip}, we employ the classification head for the [CLS] token in the image encoder of CLIP to linearly map $\mathbf{v}_j$ to $\mathbf{v}_j^\prime$ for measuring patch-text similarity, resulting in $\mathcal{V}^{\prime}_I=\{{\mathbf{v}}^{\prime}_j\}^{N_I}_{j=1}$. Therefore, a patch-wise affinity mapping $\mathcal{A} \in \mathbb{R}^{N_I \times K}$ can be computed as:
\begin{equation}
a_{jk} = CosSim ({\mathbf{v}}_j^\prime, {\bf{c}}_k) = \frac{{\mathbf{v}}_j^\prime \cdot {\mathbf{c}}_k} {\| {\mathbf{v}}_j^\prime \| \| {\mathbf{c}}_k \|},
\end{equation}
where $a_{jk}$ is the similarity score between the $j$-th local patch and the $k$-th salient concept. Then, we softly aggregate the visual concept features in ${\mathbf{C}}_s$ for each local patch $\mathbf{v}_j^\prime$ depending on $\mathcal{A}$, leading to a textual feature map $\mathcal{V}_T=\{{\mathbf{v}}^t_j\}^{N_I}_{j=1}$ for each synthetic image $I$:
\begin{equation}
{\mathbf{v}}^t_j = \sum_{k=1}^{K} \alpha_{jk} * {\bf{c}}_k =  \sum_{k=1}^{K} \frac{exp(a_{jk} / \tau)} {\sum_{m=1}^{K} exp(a_{jm} / \tau)} * {\bf{c}}_k,
\end{equation}
where $\tau$ is the temperature.

It can be observed that $\mathcal{V}_T$ only contains the semantics of salient concepts, which would constrain the captioning model to produce short captions. To alleviate the information loss, we further mix up $\mathcal{V}_I$ and $\mathcal{V}_T$ into $\mathcal{V}_{IT}$ as per $\mathcal{A}$. Technically, we pinpoint the top-$M$ visual patches in $\mathcal{V}_I$ that are highly related to the salient concept features ${\mathbf{C}}_s$ based on $\mathcal{A}$, and replace them with the corresponding ones from $\mathcal{V}_T$. In contrast to the training process with synthetic data, we concatenate the aforementioned top-$M$ patches from $\mathcal{V}_T$ with $\mathcal{V}_I$ to derive $\mathcal{V}_{IT}$ for the real images during inference. The mixed-up feature map $\mathcal{V}_{IT}$ is then passed through an encoder-decoder framework for zero-shot image captioning.

\subsection{Image Captioner}
The image captioning model is framed as a typical Transformer-based encoder-decoder framework with minor adaptation for the mixed-up feature obtained in the proposed PCM. Formally, given both the global image feature $\mathbf{g}_I$ and the mixed-up feature map $\mathcal{V}_{IT}$ for the input synthetic image $I$, we first project both these features into a joint embedding space and concatenate the results into $\mathcal{V}^{(0)}=\{{\mathbf{g}}_I^{(0)}, {\mathbf{v}}_1^{(0)}, {\mathbf{v}}_2^{(0)}, ..., {\mathbf{v}}_{N_I}^{(0)}\}$. Additionally, the position encodings and token-type embeddings are added to $\mathcal{V}^{(0)}$ following general practices, which is, in turn, fed into the encoder-decoder for caption generation.

\textbf{Visual-Semantic Encoder.}
The visual-semantic encoder consists of $N_e$ Transformer-based blocks, where each block includes a multi-head self-attention layer followed by a feed-forward layer. Formally, the output from a vanilla multi-head self-attention layer (MHA) with $H$ attention heads can be computed as:
\begin{equation}
\begin{aligned}
&{{\bf{MHA}}({\bf{Q}},{\bf{K}},{\bf{V}}) = {\bf{Concat}}({head}_{1},...,{head}_{H}){W^O}},\\
&{{head}_{i} = {\bf{Attention}}({\bf{Q}}W_i^Q,{\bf{K}}W_i^K,{\bf{V}}W_i^V)},\\
&{{\bf{Attention}}({\bf{Q}},{\bf{K}},{\bf{V}}) = {\bf{softmax}}(\frac{{{\bf{Q}}{{\bf{K}}^T}}}{{\sqrt d }}){\bf{V}}},
\end{aligned}
\end{equation}
where ${\mathbf{Concat}}( \cdot )$ is the concatenation operation. $W_i^Q$, $W_i^K$, $W_i^V$, $W^O$ are the weight matrices of the $i$-th head, and $d$ is a scaling factor. Therefore, the operation of the $i$-th block in the visual-semantic encoder can be expressed as:
\begin{equation}
\begin{aligned}
&{{\mathcal{V}^{(i + 1)}} = {\mathcal{F}}({\bf{LN}}({\mathcal{V}^{(i)}} + {\bf{MHA}}({\mathcal{V}^{(i)}},{\mathcal{V}^{(i)}},{\mathcal{V}^{(i)}})))},\\
&{{\mathcal{F}}(x) = {\bf{LN}}(x + {\bf{FC}}(\delta ({\bf{FC}}(x))))},\\
\end{aligned}
\end{equation}
where ${\bf{FC}}$, $\bf{LN}$ and $\delta$ are the fully-connected layer, layer normalization, and activation function, respectively.
Moreover, inter-layer global feature interaction is devised to obtain a more comprehensive global feature with the outputs from all the blocks as:
\begin{equation}
{\tilde{\bf{g}}}_I = W_g * \mathbf{Concat}({\bf{g}}_I^{(0)}, {\bf{g}}_I^{(1)}, ..., {\bf{g}}_I^{(N_e)} ),
\end{equation}
where $W_g$ is a learnable weight matrix. Finally, $\mathcal{V}_{IT}$ is encoded into:
\begin{equation}
\mathcal{\tilde V} = \mathbf{Concat} ({\tilde{\bf{g}}}_I, {\bf{v}}_1^{(N_e)}, {\bf{v}}_2^{(N_e)}, ..., {\bf{v}}_{N_I}^{(N_e)}).
\end{equation}

\textbf{Sentence Decoder.}
Given the fine-grained visual features $\mathcal{\tilde V}$ from the visual-semantic encoder, a Transformer-based sentence decoder is exploited to generate sentences. Similarly, the sentence decoder is implemented as $N_d$ Transformer blocks, where each Transformer-style block consists of a masked multi-head self-attention layer \cite{vaswani2017attention}, a multi-head cross-attention layer, and a feed-forward layer. Let ${\mathcal{H}}^{(0)}_{0:N_s-1}=\{ {\bf{h}}^{(0)}_0, {\bf{h}}^{(0)}_1, ..., {\bf{h}}^{(0)}_{N_s-1} \}$ stand for the textual features of the sentence that describes the input image $I$, where ${\bf{h}}^{(0)}_t$ is the textual feature of the $t$-th word $w_t$ in sentence $S$. Specifically, at the $t$-th decoding timestep, the $i$-th decoder block operates as:
\begin{equation}
\begin{aligned}
&{{\bf{h}}^{(i+1)}_t = {\mathcal{F}}({\bf{LN}}({\tilde{\bf{h}}}^{(i)}_t + {\bf{MHA}}({\tilde{\bf{h}}}^{(i)}_t,{\mathcal{\tilde V}},{\mathcal{\tilde V}})))},\\
&{{\tilde{\bf{h}}}^{(i)}_t = {\bf{LN}}({\bf{h}}^{(i)}_t + {\bf{MaskedMHA}}({\bf{h}}^{(i)}_t,{\mathcal{H}}^{(i)}_{0:t},{\mathcal{H}}^{(i)}_{0:t}))},
\end{aligned}
\end{equation}
where ${\bf{MaskedMHA}}$ is the masked multi-head self-attention layer. Finally, we use the output of final block ${\bf{h}}^{(N_d)}_t$ to predict next word $w_{t+1}$ through softmax.

\subsection{CLIP-weighted Cross-Entropy Loss}
Following the conventional training strategy in image captioning, the captioning model parametrized as $\theta$ is optimized by maximizing the likelihood of the ground-truth sentence (i.e., standard cross-entropy loss):
\begin{equation}
{\theta ^*} = \arg \,\mathop {max}\limits_\theta  \sum\limits_{(I, S)} {\log \,p(S|I; \theta )}.
\end{equation}
By applying the chain rule, the log probability of the sentence can be decomposed into the sum of the log probabilities over the ground-truth words:
\begin{equation}
\log {\mkern 1mu} p(S|I;\theta ) = \sum\limits_{t = 1}^{{N_s}} {\log \,p({w_t}|I,\{{w_i}\}^{t-1}_{i=0}; \theta )}.
\end{equation}

However, due to the existing distribution discrepancy between synthetic and real images, the standard cross-entropy loss is not optimal in this scenario, where it treats low-quality synthetic images as important as high-quality ones, thereby impeding cross-modal learning of the captioning model.
To address this challenge, we propose a novel CLIP-weighted cross-entropy loss. Unlike the standard cross-entropy loss, which treats all synthetic image-text pairs equally, our proposed CLIP-weighted cross-entropy loss dynamically adjusts the training loss for each sample based on the global vision-language similarity score \cite{hessel2021clipscore}:
\begin{equation}
\begin{aligned}
&\log {\mkern 1mu} p(S|I;\theta ) = c_I \sum\limits_{t = 1}^{{N_s}} {\log \,p({w_t}|I,\{{w_i}\}^{t-1}_{i=0}; \theta )}, \\
&c_I = CLIPScore(I, S) = min(1.0,  w * \frac{{\bf{g}}_I \cdot {\bf{g}}_S} {\| {\bf{g}}_I \| \| {\bf{g}}_S \|}),
\end{aligned}
\end{equation}
where $w$ is the scaling factor, ${\mathbf{g}}_I$ and ${\mathbf{g}}_S$ are the global CLIP embeddings of the synthetic image and the ground-truth sentence, respectively. With this additional weighting coefficient, CLIP-weighted cross-entropy loss penalizes the low-quality synthetic pairs for improved learning of vision-language alignment.

\section{Experiments}
\label{sec:experiments}

\subsection{Datasets and Experimental Settings}
\textbf{Datasets.} We empirically verify and analyze the effectiveness of our PCM-Net on two widely adopted image captioning benchmarks: MSCOCO \cite{Lin:ECCV14} and Flickr30k \cite{plummer2015flickr30k}. There are five human-annotated sentences per image in these datasets. For fair comparisons, we follow the Karpathy split \cite{karpathy2015deep} and take 113,287 images for training, 5,000 images for validation, and 5,000 images for testing on MSCOCO. For Flickr30k which consists of 31,783 images, we use 1,000 images for validation, 1,000 for testing and the rest for training. It is worth noting that we only use the text corpus in training split following the zero-shot image captioning settings. We use Stable Diffusion \cite{rombach2022high} to synthesize an image for each caption in training corpus, resulting in SynthImgCap dataset consisting of 542,401 and 144,541 synthetic image-text pairs for MSCOCO and Flickr30k, respectively.

\textbf{Implementation Details.} In PCM-Net, the visual-semantic encoder and sentence decoder are built with $N_e = N_d = 3$ Transformer blocks. The size of hidden state in each Transformer block is 512. We utilize CLIP-ViT-L/14 to extract visual and textual features as in CgT-GAN \cite{yu2023cgt}. As a result, each input image is represented as a 768-dimensional global feature vector plus a 1,024-dimensional grid feature map for local patches. Gaussian noise is injected into global feature vector in training following CapDec \cite{nukrai2022text}. We build the visual concept vocabulary $\mathcal{C}$ from the high-frequency nouns in MSCOCO training corpus, and each concept is represented as a 768-dimensional textual feature in CLIP space. The visual and textual patch features extracted by CLIP are further mapped into the common space with 512 dimensions through a fully connected layer. During the training stage, PCM-Net is optimized with Adam \cite{kingma2014adam} optimizer on CXE loss. The whole optimization process takes 12 epochs with the learning rate scheduling strategy in \cite{vaswani2017attention}. The warmup steps and batch size are set as 10,000 and 32. At inference, we adopt beam search strategy with the beam size as 3. We report the performances of PCM-Net over five evaluation metrics: BLEU-4 \cite{Papineni:ACL02} (B4), METEOR \cite{Banerjee:ACL05} (M), ROUGE \cite{lin2004rouge} (R), CIDEr \cite{vedantam2015cider} (C), and SPICE \cite{spice2016} (S).

\begin{table*}[!tb]
\centering
\setlength\tabcolsep{1.8pt}
\caption{Performance of our PCM-Net and other state-of-the-art approaches on the test split of the MSCOCO and Flickr30k datasets under the in-domain zero-shot setting. $\dag$ denotes the use of real images in training.}
\begin{tabular}{c|c|ccccc|ccccc}
\Xhline{2\arrayrulewidth}
\multicolumn{1}{c|}{\multirow{2}{*}{{Methods}}}  & \multicolumn{1}{c|}{\multirow{2}{*}{{Backbone}}}  & \multicolumn{5}{c|}{MSCOCO}       & \multicolumn{5}{c}{Flickr30k}    \\
                                        &          & B4   & M    & R    & C     & S    & B4   & M    & R    & C    & S    \\ \hline\hline
ZeroCap \cite{tewel2022zerocap}         & ViT-B/32 & 7.0  & 15.4 & 31.8 & 34.5  & 9.2  & 5.4  & 11.8 & 27.3 & 16.8 & 6.2  \\
MAGIC \cite{su2022language}             & ViT-B/32 & 12.9 & 17.4 & 39.9 & 49.3  & 11.3 & 6.4  & 13.1 & 31.6 & 20.4 & 7.1  \\
CapDec \cite{nukrai2022text}            & RN50x4   & 26.4 & 25.1 & 51.8 & 91.8  & -    & 17.7 & 20.0 & 43.9 & 39.1 & -    \\
DeCap \cite{li2023decap}                & ViT-B/32 & 24.7 & 25.0 & -    & 91.2  & 18.7 & 21.2 & 21.8 & -    & 56.7 & 15.2 \\
ViECap \cite{fei2023transferable}       & ViT-B/32 & 27.2 & 24.8 & -    & 92.9  & 18.2 & 21.4 & 20.1 & -    & 47.9 & 13.6 \\
SynTIC \cite{liu2023improving}          & ViT-B/32 & 29.9 & 25.8 & 53.2 & 101.1 & 19.3 & 22.3 & 22.4 & 47.3 & 56.6 & 16.6 \\
PCM-Net                                 & ViT-B/32 & \textbf{31.5} & \textbf{25.9} & \textbf{53.9} & \textbf{103.8} & \textbf{19.7} & \textbf{26.9}    & \textbf{23.0}    & \textbf{50.1}    & \textbf{61.3}    & \textbf{16.8}    \\ \hline
CgT-GAN \textdagger \cite{yu2023cgt}    & ViT-L/14 & 30.3 & \textbf{26.9} & 54.5 & 108.1 & 20.5 & 24.1 & 22.6 & 48.2 & 64.9 & 16.1 \\
PCM-Net                                 & ViT-L/14 & \textbf{33.6} & \textbf{26.9} & \textbf{55.4} & \textbf{113.6} & \textbf{20.8} & \textbf{28.5} & \textbf{24.3} & \textbf{51.4} & \textbf{69.5} & \textbf{18.2} \\ \Xhline{2\arrayrulewidth}
\end{tabular}
\label{tab:in_domain}
\end{table*}

\subsection{Performance Comparison}
\textbf{In-domain Zero-shot Image Captioning.} We conduct experiments on MS-COCO and Flickr30k datasets under the in-domain settings, where the model is trained on the training corpus and evaluated on the test split of the same dataset. Table \ref{tab:in_domain} summarizes the performance comparisons between the state-of-the-art models and our PCM-Net. All runs are briefly grouped into three directions: (1) zero-shot methods without further training (e.g., ZeroCap \cite{tewel2022zerocap}) that repurpose the text-to-image matching models to generate captions; (2) traditional methods (e.g., MAGIC \cite{su2022language}, CapDec \cite{nukrai2022text}, DeCap \cite{li2023decap}, ViECap \cite{fei2023transferable}) that exploit the potent cross-modal alignment capabilities of CLIP to bridge the modality gap; (3) the approaches (e.g., SynTIC \cite{liu2023improving}) that utilize synthetic image-text pairs created by off-the-shelf text-to-image generation models. As shown in the table, our PCM-Net consistently exhibits better performances than the state-of-the-art methods across all the metrics on both MSCOCO and Flickr30k datasets. In particular, the CIDEr score of PCM-Net can achieve 113.6\%, which leads to an absolute improvement of 5.5\% over CgT-GAN (CIDEr: 108.1\%). This demonstrates the effectiveness of leveraging synthetic image-text data for mitigating the modality gap in pixel space. Compared to the method that does not involve further training (e.g., ZeroCap), traditional methods (e.g., ViECap \cite{fei2023transferable}) improve the performances by exploiting the cross-modal alignment of CLIP to bridge the modality gap. 
Though SynTIC \cite{liu2023improving} manages to enhance performances by training on synthetic image-text pairs in a similar spirit, our PCM-Net substantially outperforms it with a significant margin. This demonstrates the merits of exploring fine-grained spatial visual features extracted from CLIP and applying patch-wise cross-modal feature mix-up to mitigate unfaithful content in synthetic images. Furthermore, PCM-Net leverages CLIP to prioritize high-quality synthetic image-text pairs. 
Similar trends are also observed in the Flickr30k dataset. This again confirms the advantage of our proposal.

\begin{table*}[!tb]
\centering
\setlength\tabcolsep{2.0pt}
\caption{Performance of our PCM-Net and other state-of-the-art approaches on the test split of the MSCOCO and Flickr30k datasets under the cross-domain zero-shot setting. $\dag$ denotes the use of real images in training.}
\begin{tabular}{c|c|ccccc|ccccc}
\Xhline{2\arrayrulewidth}
\multicolumn{1}{c|}{\multirow{2}{*}{{Methods}}}  & \multicolumn{1}{c|}{\multirow{2}{*}{{Backbone}}}  & \multicolumn{5}{c|}{MSCOCO $\Rightarrow$ Flickr30k} & \multicolumn{5}{c}{Flickr30k $\Rightarrow$ MSCOCO} \\
                                            &          & B4   & M    & R    & C    & S    & B4   & M    & R    & C    & S    \\ \hline\hline
MAGIC \cite{su2022language}                 & ViT-B/32 & 6.2  & 12.2 & 31.3 & 17.5 & -    & 5.2  & 12.5 & 30.7 & 18.3 & -    \\
CapDec \cite{nukrai2022text}                & RN50x4   & 17.3 & 18.6 & 42.7 & 35.7 & -    & 9.2  & 16.3 & 36.7 & 27.3 & -    \\
DeCap \cite{li2023decap}                    & ViT-B/32 & 16.3 & 17.9 & -    & 35.7 & 11.1 & 12.1 & 18.0 & -    & 44.4 & 10.9 \\
ViECap \cite{fei2023transferable}           & ViT-B/32 & 17.4 & 18.0 & -    & 38.4 & 11.2 & 12.6 & 19.3 & -    & 54.2 & 12.5 \\
SynTIC \cite{liu2023improving}              & ViT-B/32 & 17.9 & 18.6 & 42.7 & 38.4 & 11.9 & 14.6 & 19.4 & 40.9 & 47.0 & 11.9 \\
PCM-Net                                     & ViT-B/32 & \textbf{20.8} & \textbf{19.2} & \textbf{45.2} & \textbf{45.5} & \textbf{12.9} & \textbf{17.1} & \textbf{19.6} & \textbf{43.2} & \textbf{54.9} & \textbf{12.8} \\ \hline
CgT-GAN \textdagger \cite{yu2023cgt}        & ViT-L/14 & 17.3 & 19.6 & 43.9 & 47.5 & 12.9 & 15.2 & 19.4 & 40.9 & 58.7 & 13.4 \\
PCM-Net                                     & ViT-L/14 & \textbf{23.9} & \textbf{21.2} & \textbf{47.8} & \textbf{55.9} & \textbf{14.2} & \textbf{17.9} & \textbf{20.3} & \textbf{44.0} & \textbf{61.3} & \textbf{13.5} \\ \Xhline{2\arrayrulewidth}
\end{tabular}
\label{tab:cross_domain}
\end{table*}

\textbf{Cross-domain Zero-shot Image Captioning.}
Next, we evaluate our PCM-Net under the cross-domain settings, where the model is evaluated on the test split of a different dataset. As shown in Table \ref{tab:cross_domain}, the performance trends in cross-domain settings are similar to those in in-domain settings. Concretely, our PCM-Net surpasses the state-of-the-art techniques (CgT-GAN) by an absolute improvement of 8.4\% in CIDEr score on the MSCOCO $\Rightarrow$ Flickr30k task. The results again demonstrate the effectiveness of unfaithful content mitigation and emphasis on high-quality synthetic pairs for zero-shot image captioning.

\begin{figure}[!tb]
    \centering {\includegraphics[width=0.98\textwidth]{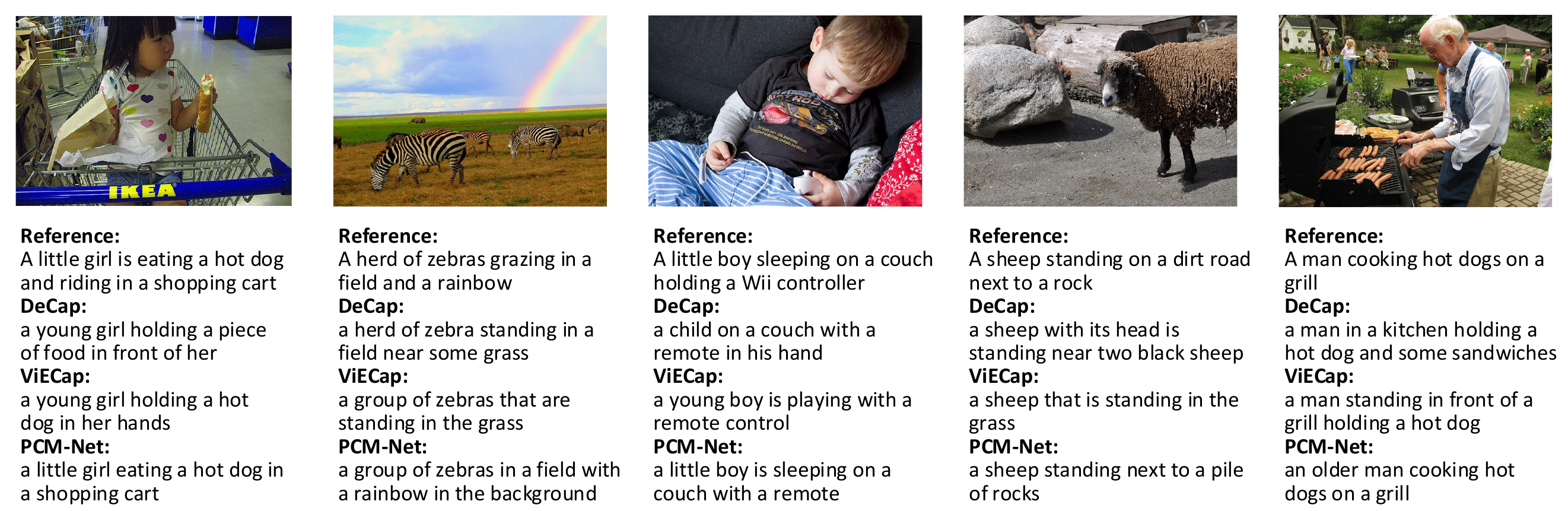}}
    \caption{Examples of image captioning results generated by DeCap \cite{li2023decap}, ViECap \cite{fei2023transferable} and our PCM-Net, coupled with the corresponding ground-truth sentences(Reference).}
    \label{fig:case}
\end{figure}

\textbf{Qualitative Analysis.} Figure \ref{fig:case} shows several qualitative results of our PCM-Net and two approaches(i.e., DeCap and ViECap) on MSCOCO dataset, coupled with one human-annotated ground-truth sentence (Reference). From these exemplar results, it is easy to see that our PCM-Net can predict more semantically relevant and logically correct sentences. For instance, in the first example, both DeCap and ViECap are only aware of the major objects (girl and hot dog), while ignoring the salient object of shopping cart. 
Instead, by employing a text-to-image diffusion prior for zero-shot image captioning and applying PCM plus CXE loss to boost the training, our PCM-Net effectively captures all significant objects in the image (girl, hot dog, and shopping cart), yielding more accurate and descriptive sentences.

\subsection{Experimental Analysis}

\begin{table}[t]
\centering
\setlength\tabcolsep{6.0pt}
\caption{Ablation study on each design in PCM-Net on MSCOCO under the in-domain zero-shot setting. \textbf{Base} denotes the vanilla Transformer-based encoder-decoder model. \textbf{Mix-up} represents the use of patch-wise cross-modal feature mix-up mechanism. \textbf{CXE} refers to the CLIP-weighted cross-entropy loss.}
\begin{tabular}{ccc|ccccc}
\Xhline{2\arrayrulewidth}
\multirow{2}{*}{\textbf{Base}} & \multirow{2}{*}{\textbf{Mix-up}} & \multirow{2}{*}{\textbf{CXE}} & \multicolumn{5}{c}{MSCOCO}        \\ 
                      &                         &                      & B4   & M    & R    & C     & S    \\ \hline \hline

$\checkmark$          & \multicolumn{1}{l}{}    & \multicolumn{1}{l|}{} & 33.0 & 26.3 & 54.8 & 108.9 & 20.2 \\
$\checkmark$          & $\checkmark$            & \multicolumn{1}{l|}{} & 33.3 & 26.8 & 55.0 & 111.6 & 20.7 \\
$\checkmark$          & \multicolumn{1}{l}{}    & $\checkmark$          & 33.2 & 26.4 & 55.1 & 112.0 & 20.5 \\
$\checkmark$          & $\checkmark$            & $\checkmark$          & \textbf{33.6} & \textbf{26.9} & \textbf{55.4} & \textbf{113.6} & \textbf{20.8} \\
\Xhline{2\arrayrulewidth}
\end{tabular}
\label{tab:ablation}
\end{table}

\begin{figure}[!tb]
    \centering {\includegraphics[width=0.95\textwidth]{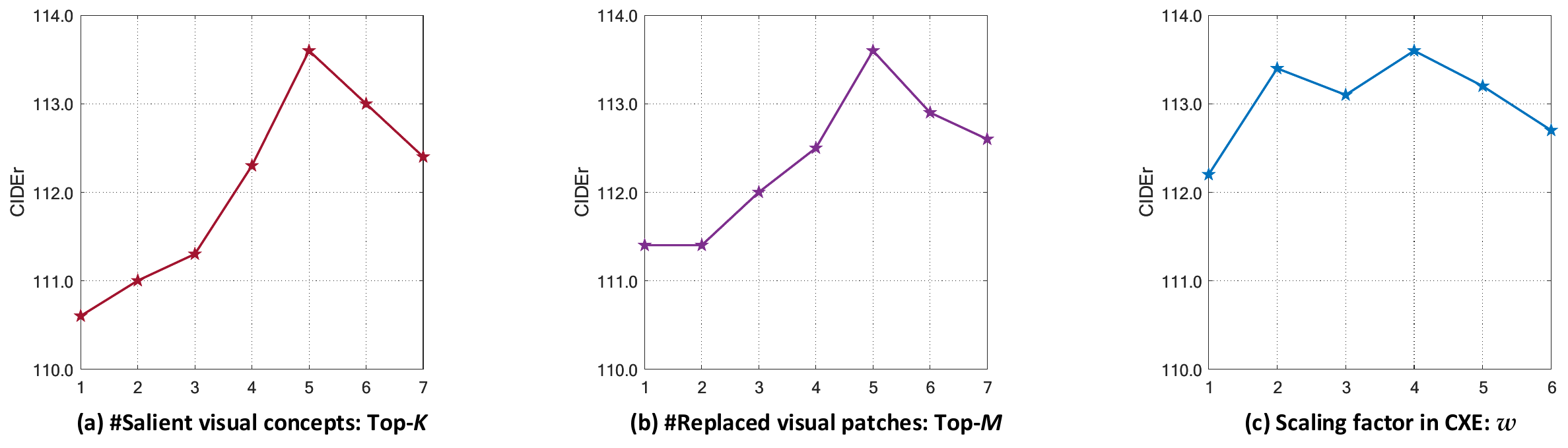}}
    \caption{Ablation study on hyperparameters in PCM-Net on MSCOCO.}
    \label{fig:ablation_hyper}
\end{figure}

\textbf{Ablation Study.} We conduct ablation study to investigate how each component in our PCM-Net influences the overall performances on MSCOCO Karpathy test split. Table \ref{tab:ablation} presents the performance comparisons among different ablated runs of our PCM-Net. We start with a Transformer-based encoder-decoder model (\textbf{Base}), which is trained on synthetic pairs in a supervised manner. Next, by incorporating the patch-wise cross-modal feature mix-up mechanism (\textbf{Base+Mix-up}) during training, we observe clear performance boosts. This implies that the mixed cross-modal features can mitigate unfaithful content in synthetic images, and thus improve visual-semantic learning. In addition, we also apply the CLIP-weighted cross-entropy loss to the base model. In this way, \textbf{Base+CXE} exhibits better performance, verifying the merit of concentration on high-quality pseudo pairs. Finally, we integrate both patch-wise cross-modal feature mix-up mechanism and the CLIP-weighted cross-entropy loss, \textbf{Base+Mix-up+CXE} (i.e., our PCM-Net) achieves the best performances across all the evaluation metrics.

\textbf{Ablation on Hyperparameters in PCM and CXE.} We perform ablation analyses on three crucial hyperparameters in our PCM-Net. 
Firstly, we investigate the impact of varying the number ($K$) of salient visual concepts in PCM. In Figure \ref{fig:ablation_hyper} (a), a consistent performance improvement is observed as $K$ increases from 1 to 5, with a slight decline beyond 5, likely due to the introduction of noisy concepts with an excessively large $K$.
Secondly, we explore the influence of replacing different numbers ($M$) of visual patches in PCM. Figure \ref{fig:ablation_hyper} (b) exhibits similar trends, indicating that excessive replacement of meaningful visual patches by text-based patches leads to a dominance of text features and subsequent degradation in representation capacity.
Lastly, we scrutinize the hyperparameter in CXE loss by varying the scaling factor ($w$) from 1 to 6. Figure \ref{fig:ablation_hyper} (c) shows the best results when $w$ is set to 4, but performance degrades when $w$ exceeds 4 due to higher CLIPScore weights being assigned to low-quality pairs, reducing CXE loss to vanilla cross-entropy loss. Based on these empirical analyses, we set $K$, $M$, and $w$ as 5, 5, and 4, respectively.

\begin{table}[t]
\centering
\setlength\tabcolsep{2.0pt}
\caption{Performance of ViECap and two variants equipped with our proposals on the validation split of NoCaps \cite{agrawal2019nocaps} benchmark under the cross-domain zero-shot setting.}
\resizebox{\linewidth}{!}{
\begin{tabular}{c|l|cccccccc}
\Xhline{2\arrayrulewidth}
\multirow{3}{*}{Methods}        & \multicolumn{1}{c|}{\multirow{3}{*}{Backbone}} & \multicolumn{8}{c}{MSCOCO $\Rightarrow$ NoCaps val}                                                                                                                                         \\ \cline{3-10} 
                                & \multicolumn{1}{c|}{}                          & \multicolumn{2}{c|}{In-domain}                     & \multicolumn{2}{c|}{Near-domain}                   & \multicolumn{2}{c|}{Out-of-domain}                & \multicolumn{2}{c}{Overall}   \\
                                & \multicolumn{1}{c|}{}                          & C             & \multicolumn{1}{c|}{S}             & C             & \multicolumn{1}{c|}{S}             & C             & \multicolumn{1}{c|}{S}            & C             & S             \\ \hline \hline
ViECap \cite{fei2023transferable}   & ViT-B/32+GPT-2$_{Base}$                        & 61.1          & \multicolumn{1}{c|}{10.4}          & 64.3          & \multicolumn{1}{c|}{9.9}           & 65.0          & \multicolumn{1}{c|}{8.6}          & 66.2          & 9.5           \\
\multicolumn{1}{l|}{SYN-ViECap} & ViT-B/32+GPT-2$_{Base}$                        & 61.7          & \multicolumn{1}{c|}{10.5}          & 68.5          & \multicolumn{1}{c|}{10.3}          & 71.4          & \multicolumn{1}{c|}{9.4}          & 70.5          & 10.0          \\
\multicolumn{1}{l|}{PCM-ViECap} & ViT-B/32+GPT-2$_{Base}$                        & \textbf{66.0} & \multicolumn{1}{c|}{\textbf{10.7}} & \textbf{72.7} & \multicolumn{1}{c|}{\textbf{10.6}} & \textbf{75.7} & \multicolumn{1}{c|}{\textbf{9.7}} & \textbf{74.7} & \textbf{10.3} \\ \Xhline{2\arrayrulewidth}
\end{tabular}
}
\label{tab:nocaps}
\end{table}

\subsection{Framework Compatibility}
To verify the generalizability of our methods, we follow the training mechanism of ViECap \cite{fei2023transferable} and evaluate the models on another widely used benchmark, NoCaps \cite{agrawal2019nocaps}. We first upgrade ViECap to SYN-ViECap by replacing the CLIP textual features of the input sentence with the CLIP visual features of the generated image by diffusion model for training. As shown in Table \ref{tab:nocaps}, SYN-ViECap outperforms ViECap across all metrics on the NoCaps validation split, demonstrating the advantage of text-to-image diffusion priors for ZIC. By further integrating PCM and CXE into SYN-ViECap, PCM-ViECap achieves a substantial performance boost across all metrics, particularly excelling in the Out-of-domain category. This indicates its effectiveness in narrowing the distribution discrepancy between synthetic and real images. In summary, our proposed PCM mechanism and CXE loss can generalize well on various model structures and achieve remarkable zero-shot transferability.

\begin{figure}[!tb]
    \centering {\includegraphics[width=0.65\textwidth]{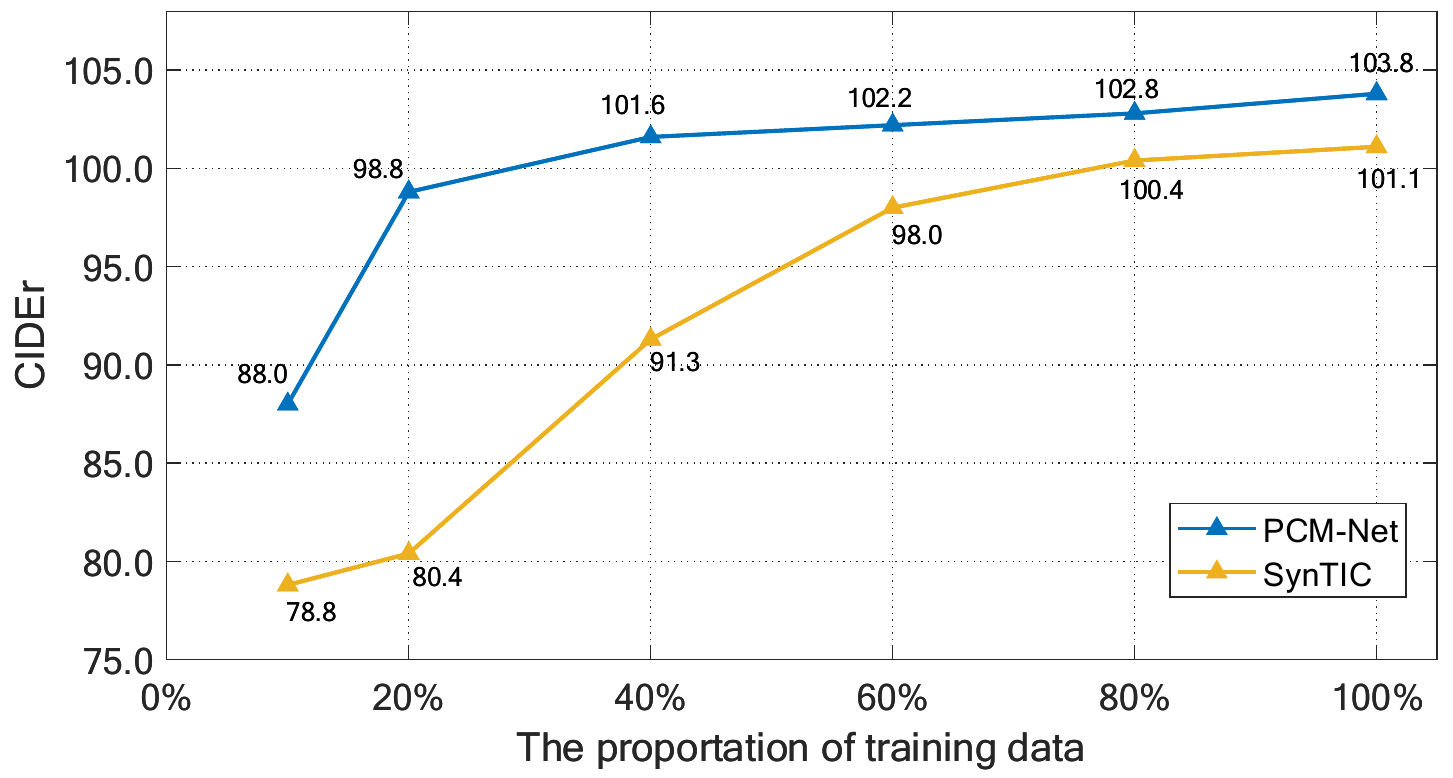}}
    \caption{Performance comparison of PCM-Net and SynTIC on MSCOCO with various proportions of training data.}
    \label{fig:ablation_data_eff}
\end{figure}

\subsection{Effect of the Training Data Size}
We also explore the effect of training data size on PCM-Net following \cite{fei2023transferable,liu2023improving}. This is done by randomly sampling various proportions of data from MSCOCO to train PCM-Net. As depicted in Figure \ref{fig:ablation_data_eff}, PCM-Net consistently outperforms SynTIC across all data scales, especially in low-data scenarios. Notably, PCM-Net trained with only 40\% data surpasses SynTIC that uses the entire training dataset (101.6 vs. 101.1). In summary, PCM-Net performs robustly with limited training data and can be further improved by increasing the training data size.

\section{Conclusion}
\label{sec:conclusion}
In this paper, we propose PCM-Net for zero-shot image captioning, which aims to mitigate the modality gap of CLIP for visual-semantic learning by leveraging synthetic image-text data for training. Particularly, we employ an off-the-shelf text-to-image diffusion model to build a synthetic dataset, i.e., SynthImgCap. For each input image, salient visual concepts are first constructed by performing zero-shot entity classification of images in CLIP space. After that, the patch-wise cross-modal feature mix-up mechanism is proposed to mix the fine-grained patch features of the synthetic image with the textual features, reducing the flawed or unfaithful details in the synthetic images. Finally, a visual-semantic encoder-decoder is exploited to refine the derived features and generate a caption. To improve the training of captioning model with noisy synthetic data, we propose a novel CLIP-weighted cross-entropy loss to prioritize the high-quality image-text pairs over the low-quality counterparts. Extensive experiments conducted on MSCOCO and Flickr30k datasets demonstrate the superiority of our PCM-Net.

\textbf{Broader Impact.}
Our PCM-Net is trained to produce image captions based on learnt statistics of the training corpus and synthetic images, and thus the biases rooted in those data will be reflected in the outputs, resulting in negative societal impacts. Hence more future research is necessary for addressing the issue.

\section*{Acknowledgments}
This work is partially supported by China NSFC under Grant No. 61772563. 

%
%
\bibliographystyle{splncs04}
\bibliography{main}

\end{document}